\def\year{2020}\relax
\documentclass[letterpaper]{article} 
\usepackage{aaai20}  
\usepackage{times}  
\usepackage{helvet} 
\usepackage{courier}  
\usepackage[hyphens]{url}  
\usepackage{graphicx} 
\usepackage{graphicx}  
\usepackage{amsmath}
\usepackage{amssymb}
\usepackage{bm}
\title{Shape-Aware Organ Segmentation by Predicting Signed Distance Maps}
\author{Yuan Xue,\textsuperscript{\rm 1}\thanks{This work was done during the first author's summer internship at Tencent Hippocrates Research Lab.} Hui Tang,\textsuperscript{\rm 2} Zhi Qiao,\textsuperscript{\rm 2} Guanzhong Gong,\textsuperscript{\rm 3} \\\Large \textbf{Yong Yin,\textsuperscript{\rm 3} Zhen Qian,\textsuperscript{\rm 2} Chao Huang,\textsuperscript{\rm 2} Wei Fan,\textsuperscript{\rm 2} Xiaolei Huang\textsuperscript{\rm 1}}\\
\textsuperscript{\rm 1}Pennsylvania State University, University Park, PA, USA\\
\textsuperscript{\rm 2}Tencent Hippocrates Research Lab, Palo Alto, CA, USA\\
\textsuperscript{\rm 3}Shandong Cancer Hospital and Institute, Jinan, Shandong, China\\ 
}
 
\begin{document}

\maketitle

\begin{abstract}
In this work, we propose to resolve the issue existing in current deep learning based organ segmentation systems that they often produce results that do not capture the overall shape of the target organ and often lack smoothness. Since there is a rigorous mapping between the Signed Distance Map (SDM) calculated from object boundary contours and the binary segmentation map, we exploit the feasibility of learning the SDM directly from medical scans. By converting the segmentation task into predicting an SDM, we show that our proposed method retains superior segmentation performance and has better smoothness and continuity in shape. To leverage the complementary information in traditional segmentation training, we introduce an approximated Heaviside function to train the model by predicting SDMs and segmentation maps simultaneously. We validate our proposed models by conducting extensive experiments on a hippocampus segmentation dataset and the public MICCAI 2015 Head and Neck Auto Segmentation Challenge dataset with multiple organs. While our carefully designed backbone 3D segmentation network improves the Dice coefficient by more than $5\%$ compared to current state-of-the-arts, the proposed model with SDM learning produces smoother segmentation results with smaller Hausdorff distance and average surface distance, thus proving the effectiveness of our method.  
\end{abstract}

\section{Introduction} 
In medical image segmentation, organ segmentation is of great importance in disease diagnosis and surgical planning. For instance, the segmented shape of hippocampus may be useful as a biomarker for neurodegenerative disorders including the Alzheimer’s disease (AD)~\cite{scher2007hippocampal}. In radiotherapy planning, accurate segmentation result of organs at risks (OARs) may help oncologists design better radiation treatment plans such as the appropriate beam paths so that radiation concentrates on the tumour region while minimises the dose to surrounding healthy organs~\cite{moore2011experience}. Since manual annotation of organs brings extra work load and can be error-prone, an automatic and accurate organ segmentation system has been desired. Traditional methods such as the atlas-based methods~\cite{aljabar2009multi} and active contour models~\cite{kass1988snakes} suffer from computational overhead during inference and may lack generality, deep learning based segmentation methods have been more prominent and enabled faster and more accurate segmentation results recently.

\begin{figure}[t]
\centering
\includegraphics[width=0.90\columnwidth]{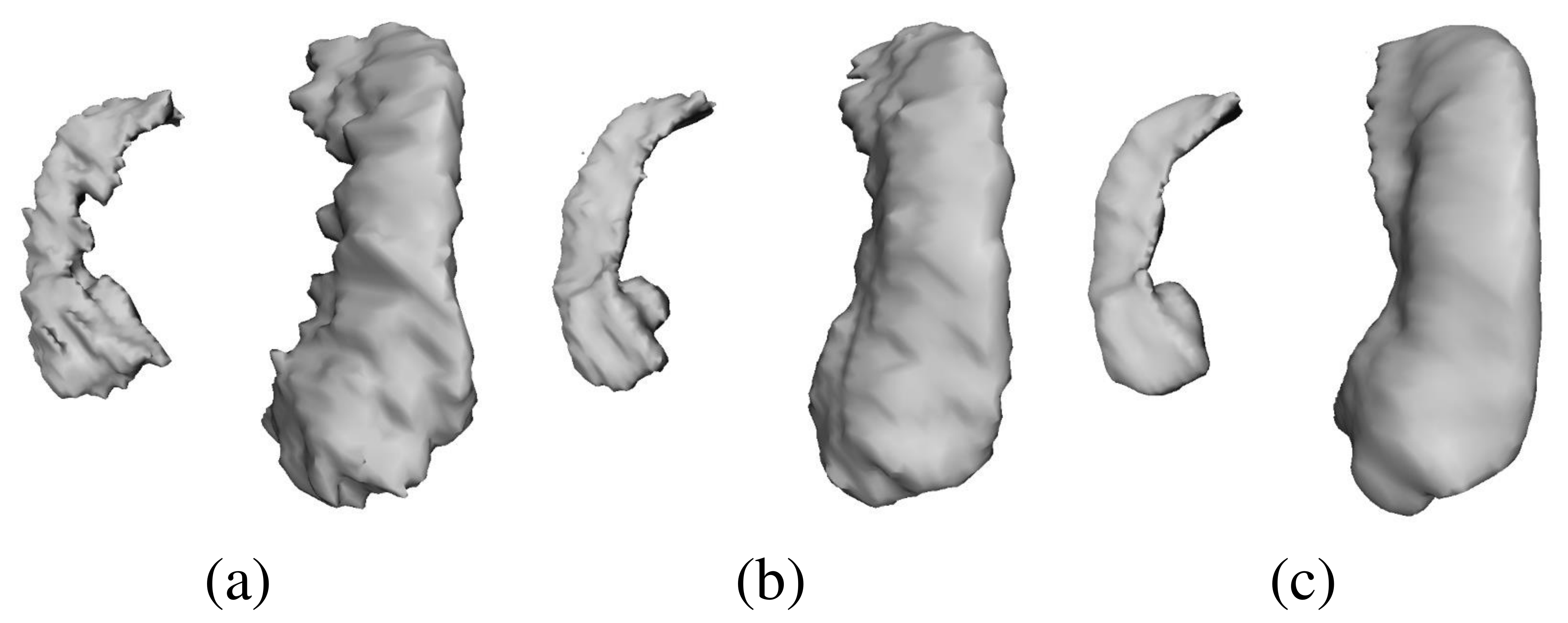} 
\caption{An example hippocampus segmentation comparison of (a) groundtruth annotation, which lacks smoothness in 3D view due to the inconsistency of annotation in 2D; (b) segmentation result from the model without predicting the signed distance map; (c) segmentation result from the model with predicting the signed distance map, and is clearly smoother while preserving the overall shape.}
\label{exp1}
\end{figure}

Different from general segmentation problems such as lesion segmentation, organs have relatively stable positions, shapes and sizes. While current state-of-the-art organ segmentation systems are dominated by deep learning based methods~\cite{roth2015deeporgan}, they often lack awareness of the feasible shape and suffer from nonsmoothness of the training ground truth labelled by doctors, especially in 3D scenarios. See Figure~\ref{exp1} as an example: the ground truth label of a hippocampus cannot maintain consistent and continuous shape due to the fact that it is annotated in 2D slices by contours instead of 3D surfaces. In traditional medical image segmentation methods, such smoothness issue can be mitigated by adding a regularization term with physical meaning as in snakes~\cite{kass1988snakes} and level sets~\cite{osher1988fronts}. To leverage the shape awareness of traditional methods, we propose to regress the Signed Distance Map (SDM) directly from the input images through a 3D convolutional neural network. 

Given a point (voxel) in image space, the absolute value of SDM is defined by the distance between the point and the closest boundary of the target organ. The sign denotes whether the point is inside the boundary of the target organ (negative) or outside the organ (positive). As an implicit shape representation, SDM embeds points and contours in a higher dimensional space, thus encodes richer information about structural features. Compared with binary segmentation map where local changes only affect local points, small changes in shape will change SDM values of multiple points globally. To predict the SDM accurately, a model has to learn the volume, position and shape information of the target organ. By enforcing the model output the global SDM of the target organ(s), we implicitly introduce continuity and smoothness terms into the segmentation process.

Enforcing the model output to be strict SDMs can prove challenging though and it should be more feasible to train a model to generate SDM and segmentation map jointly, since there is a rigorous mapping between them. 
While calculating the SDM given a segmentation map is non-trivial~\cite{zhao2005fast}, converting from SDM to segmentation map is simply by applying a Heaviside step function, where negative points are labeled as the organ region and positive points are labeled as background region.  Thus we can leverage state-of-the-art segmentation methods that produce high-quality segmentation maps and incorporate SDM learning to further enforce smoothness and shape priors.

The main contributions of this work are summarized as follows: (1) We propose a 3D UNet~\cite{cciccek20163d} based 3D segmentation backbone network with large receptive fields. We carefully design the model architecture to contain deeper layers and make it perform well on segmenting both large and small organs. Our fine-tuned backbone network achieves state-of-the-art results on the MICCAI 2015 Head
and Neck Auto Segmentation Challenge dataset~\cite{raudaschl2017evaluation}. (2) We incorporate our newly proposed SDM learning mechanism into the backbone network. To the best of our knowledge, this is the first time that the SDM is predicted in conjunction with the segmentation map instead of being a regularizer in organ segmentation tasks. The two outputs are connected through a differentiable Heaviside function and trained jointly. We also introduce a new regression loss which leads to larger gradient magnitudes for inaccurate predictions and shows better performances compared with $L_1$ regression loss in ablation studies. (3) We conduct extensive experiments on both a single organ hippocampus CT segmentation dataset and the public MICCAI 2015 Head and Neck dataset with multiple organs. Our methods outperform previous state-of-the-arts in all evaluation metrics. The segmentation maps converted from the predicted SDMs clearly show better shape and smoothness attributes than results generated by models trained without SDM.

\section{Related Works}
\subsection{Organ Segmentation}
For organ segmentation, traditional methods include statistical models~\cite{cerrolaza2015automatic}, atlas-based methods~\cite{aljabar2009multi}, active contour models~\cite{kass1988snakes} and level sets~\cite{osher1988fronts}. The performances of atlas-based methods often rely on the accuracy of registration and label fusion algorithms; Snakes and level sets require iterative optimization through gradient descent during the inference. On the contrary, advances in deep learning based 2D~\cite{ronneberger2015u} and 3D~\cite{cciccek20163d} segmentation methods have enabled more accurate organ segmentation.

Although learning based methods have faster inference speed and higher accuracy than traditional methods, they often lack awareness of the shape of the target organ. Regardless of the choice for network architecture and training loss, the segmentation output may contain extraneous regions and may not preserve the anatomical shape of the organ. Therefore, post-processing is required for error correction to refine the segmentation results.  \cite{kamnitsas2017efficient} uses fully connected conditional random fields (CRF) as the post-processing step to refine the initial segmentation result. \cite{kohlberger2011automatic} initializes the multi-region level set segmentation with result from a learning based method. They propose multiple level set constrains including a smoothness term to refine the initial result. \cite{gibson2018automatic} improves the 3D CNN segmentation result by removing small isolated regions and employing curvature flow smoothing.  

Post-processing of segmentation results can remedy the defects of learning based organ segmentation methods to some extent, especially in terms of shape awareness and smoothness. However, an end-to-end model that can automatically produce satisfying segmentation maps without the need of post-processing is more desirable.
\cite{tang2018integrating} integrates the level set smoothness term into the training process by first training with Dice loss only, then fine tuning with the Dice loss and level set term jointly. \cite{xue2018segan} incorporates adversarial learning into the segmentation network to capture both global and local image features and generate smoother segmentation maps. Although they achieve promising results, their learning targets are still binary masks which lack global shape representation.

Recent works on medical image segmentation have focused more on delineating the boundaries of target organs or lesions since the surface can be regarded as a representation of shape information. \cite{kervadec2019boundary} proposes a boundary loss which integrates over the boundary between regions to mitigate the highly unbalanced segmentation issue. \cite{ni2018elastic} converts the 3D segmentation problem into a 2D surface prediction problem by building an elastic shell and converging it to the boundary of the target organ. They achieve comparable performance to both 2D and 3D competitors with the Elastic Boundary Projection (EBP) algorithm. Different from recently proposed boundary based segmentation methods, learning a global SDM directly has the advantage of capturing better spatial relationship between voxels and providing a confidence map of the segmentation result.

\subsection{Signed Distance Map}
Several works have explored the applications of SDM or Signed Distance Function (SDF) in computer vision and graphics. \cite{perera2015motion} uses truncated SDF to better reconstruct volumetric surfaces on RGB-D images. \cite{hu2017deep} treats the linearly shifted saliency map as the SDF and refines the predicted saliency map in multiple training stages with level set smoothness terms. 
\cite{park2019deepsdf} learns the continuous 3D SDF directly from point samples by a network containing series of fully connected layers and a $L_1$ regression loss. The learned SDFs are used for obtaining state-of-the-art shape representation and completion results. 

Since medical images contain richer contextual information than point samples, more sophisticated network architecture and training strategy need to be considered when applying SDM learning on organ segmentation tasks. \cite{al2018spnet} proposes to use unsigned distance map as an intermediate step for 2D organ shape prediction. The conversion from distance map to shape parameter vector is done by PCA and the segmentation map is not involved. For the higher-dimensional 3D organ segmentation task, directly applying their method may not work well in small organs. Recently,  \cite{dangi2019distance} and~\cite{navarro2019shape} use distance map prediction as a regularizer during training for organ segmentation. Since they predict segmentation map and distance map in different branches, correspondences between predictions of the separate segmentation and SDM branches are not guaranteed. Thus, their method differs from ours in which the segmentation map and SDM are connected by a differentiable Heaviside function and can be predicted as a whole.

\section{Methodology}
\begin{figure}[t]
\centering
\includegraphics[width=0.95\columnwidth]{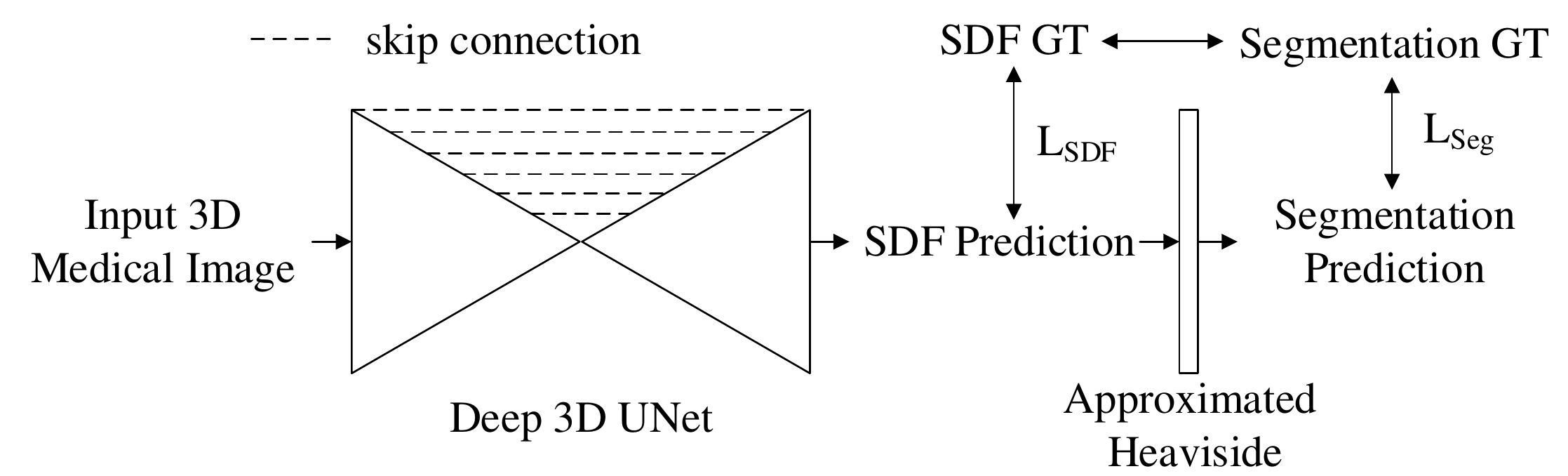} 
\caption{Illustration of our proposed SDM learning model for organ segmentation. During training, we use the differentiable approximated Heaviside function to train the proposed backbone deep 3D UNet by SDM loss and segmentation loss. }
\label{fig_method}
\end{figure}

In this section, we introduce our 3D segmentation network and the proposed SDM learning model. The overall pipeline of our proposed method is demonstrated in Figure~\ref{fig_method}. The segmentation network takes the whole 3D medical scans as input. Conventional deep learning based segmentation networks are trained by supervision from the groundtruth segmentation maps. Since there is a rigorous mapping between segmentation map and SDM, we propose to incorporate the SDM learning into the current segmentation model. The groundtruth SDM is calculated based on the groundtruth segmentation map, and any 3D segmentation network can be fit into our proposed SDM prediction and segmentation model with nearly no additional overhead. However, the SDM prediction of organs with various shape, size and location is a non-trivial problem. Thus, the model architecture and training strategy need to be carefully designed to achieve satisfactory results. Next we first introduce the architecture design of the backbone segmentation network, then introduce our SDM learning model. 

\subsection{Deep 3D UNet}
Considering the SDM is a global mapping from the 3D image space, we choose to use 3D inputs to better capture the overall organ shape in 3D space and provide more global features to the model. Our backbone segmentation network is adapted from the widely used 3D UNet~\cite{cciccek20163d}. We make several major changes to the original 3D UNet architecture and validate our proposed backbone network through comprehensive experiments on two datasets as described in the Experiments section.

One of the major challenges of organ segmentation, especially multi-organ segmentation is that the organ sizes are often highly unbalanced. The accurate segmentation of small organs remains to be an active research topic. In~\cite{zhu2019anatomynet} and~\cite{gao2019focusnet}, authors claim that having multiple downsampling operations loses high resolution information and degrades the learning performance of small organs. To this end, they both use a 3D segmentation network with only 1 downsampling operation. However, having a small receptive field results in local features and makes the model lacking in awareness of spatial relationship between far-away voxels. For SDM prediction, we expect that the model can benefit from larger receptive fields or more downsampling layers. Thus, we experiment with 3D UNet with more downsampling layers rather than less.

We finetune the architecture and model hyperparameters on the aforementioned MICCAI 2015 dataset. The best result is achieved by a 3D UNet variant with 6 downsampling operations, meaning that the largest receptive field has size $64^3$. The result is consistent for both training with traditional segmentation output and the joint training along with the SDM prediction. We argue that although features with only large receptive field can degrade the segmentation result for small organs, 3D UNet-like architecture actually learns multi-scale features with mixed receptive field sizes by skip connections. As shown in Figure~\ref{fig_method}, our final model contains 6 skip connections between different scales of feature maps. We also replace the relu activation with leaky-relu activation, and replace the deconvolution with trilinear upsampling followed by convolution. More importantly, we use group normalization~\cite{wu2018group} instead of batch normalization as in previous works. The group normalization is designed for training with smaller batchsizes. In 3D organ segmentation, input image size is often much larger than in 2D segmentation which leads to smaller training batchsizes due to GPU memory limitations. Thus, group normalization is more appropriate than batch normalization in 3D segmentation. By carefully designing and adjusting model architecture, our backbone model with more downsampling and upsampling operations outperforms the model with only 1 downsampling layer significantly for both large and small organs. More details are discussed in the Experiments section.

For our backbone network training with only segmentation output, we choose the Dice loss in all experiments. The Dice loss measures the overlapping between groundtruth and predicted segmentation maps and is defined as:
\begin{small}
\begin{equation}
\mathcal{L}_\text{Seg}=\mathcal{L}_\text{Dice}=N-\sum_{t=1}^{N}  2 \frac{\sum y_{t} p_{t} + \epsilon}{\sum y_{t}+\sum p_{t} + \epsilon} \enspace, \label{eq:dice}
\end{equation}
\end{small}
where $N$ is the number of classes, t denotes the $t$-th organ class. $y_t$ and $p_t$ represent the groundtruth annotation and model prediction, respectively. $\epsilon$ is a term with small value to avoid numerical issues.

\subsection{SDM Learning}
\begin{figure}[t]
\centering
\includegraphics[width=0.95\columnwidth]{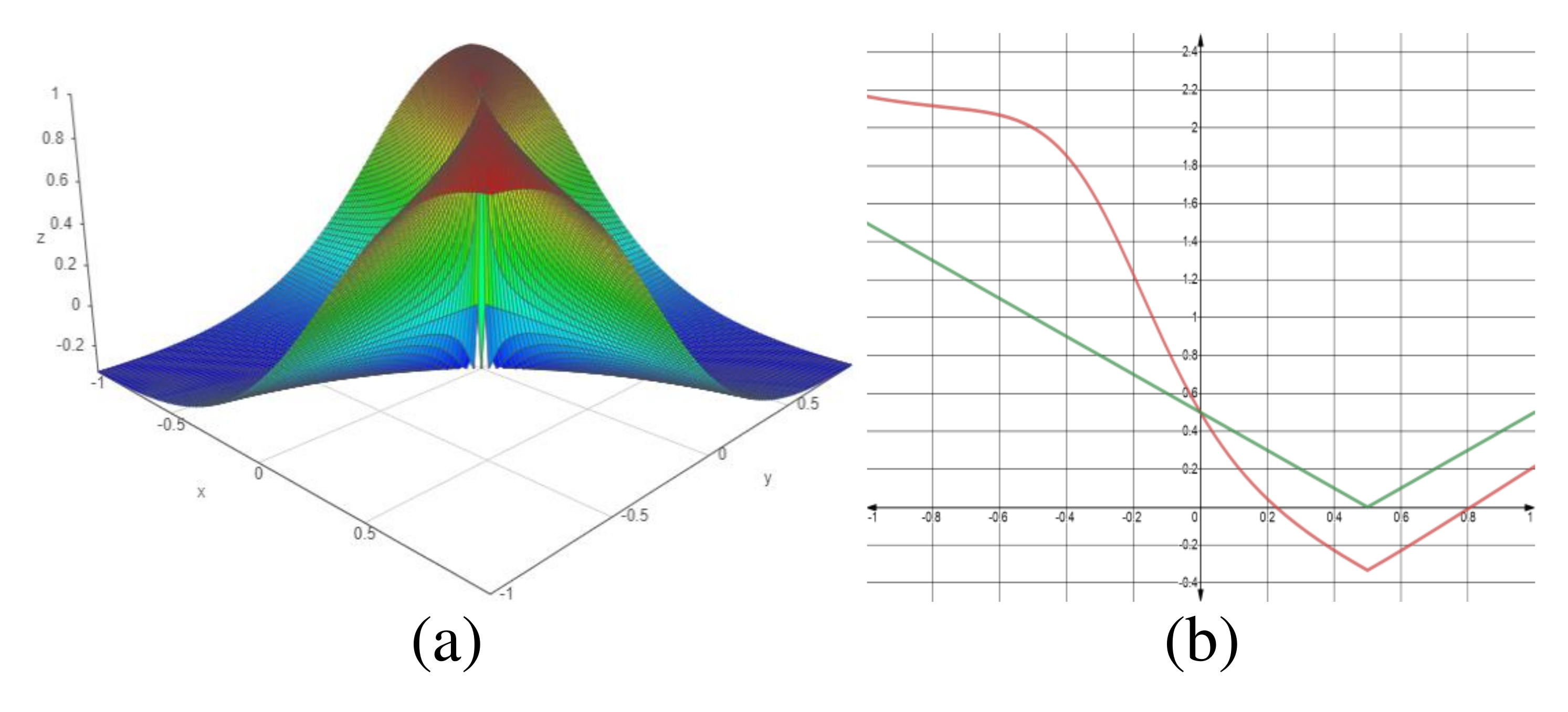} 
\caption{(a) The proposed regression loss for SDM prediction. All SDM values are normalized. (b) Plot of the loss value given the groundtruth SDM value is $0.5$. Red curve represents the combination of our proposed loss and the $L_1$ loss, green curve represents the $L_1$ loss.}
\label{fig_loss}
\end{figure}
Given a target organ and a point $x$ in the 3D medical image, the Signed Distance Map (SDM) which maps $\mathbb{R}^3$ to $\mathbb{R}$ is defined as:
\begin{small}
\begin{equation}
\phi(x)=\left\{\begin{aligned} 0,\enspace& x \in \mathcal{S} \\-\inf _{y \in \mathcal{S}}\|x-y\|_{2},\enspace& x \in \Omega_{\text{in}} \\+\inf _{y \in \mathcal{S}}\|x-y\|_{2},\enspace& x \in \Omega_{\text{out}} \end{aligned}\right.
\end{equation}
\end{small}
where $\mathcal{S}$ represents the surface of the target organ, $\Omega_{\text{in}}$ and $\Omega_{\text{out}}$ denote the region inside and outside the target organ, respectively. In other words, the absolute value of SDM indicates the distance from the point to the closest point on organ surface, while the sign indicates either inside or outside the organ. Note that the zero distance or zero level set means that the point is on the surface of the organ.

In this work, we approximate the groundtruth SDM using Danielsson's algorithm~\cite{danielsson1980euclidean} based on the groundtruth segmentation map. The approximated SDMs are used as groundtruth for training the SDM prediction model. Since input images have various fields of view and organ volumes, we further normalize the $\phi(x)$ to be in the range $[-1,1]$ for each input image and use the tanh activation in the output layer. The normalization is done by dividing the SDM by the maximum positive value for points outside the organ, or by the minimum negative value for points inside the organ. 

We started our experiments with predicting SDM and segmentation map separately in two independent branches as in~\cite{audebert2019distance}, \cite{dangi2019distance} and \cite{navarro2019shape}. However, we could not guarantee the correspondence between outputs of the two branches and thus could not let them help each other. During inference, such correspondence can be guaranteed by passing the SDM of each organ through a Heaviside step function to generate the segmentation map. Unfortunately, the Heaviside function is not differentiable and thus cannot be included in training. To match the distance map output with the segmentation output and make all network layers differentiable, we propose to use a smooth approximation to the Heaviside function which is defined as
\begin{small}
\begin{equation}
f(z) = \frac{1}{1+e^{-z/k}} \enspace,\label{eq:logistic}
\end{equation}
\end{small}
where $k$ controls the steepness of the curve and closeness to the original Heaviside function, larger $k$ means closer approximation. In our experiments, we set the $k$ to be $ 1500$ for normalized SDM due to the fact that such value guarantees around $99.9\%$ overlapping between converted segmentation map and the original segmentation map.

In~\cite{park2019deepsdf}, the authors use clamped $L_1$ loss which is the $L_1$ difference between the predicted and real SDM values. Although $L_1$ loss is robust to outliers, for multi-organ segmentation tasks, training by $L_1$ loss sometimes leads to unstable training process. Ablation results regarding the $L_1$ loss training of SDM can be found in the Experiments section. To overcome the shortcoming of $L_1$ loss, we combine the $L_1$ loss with our proposed regression loss based on a product that is defined as:
\begin{small}
\begin{equation}
\mathcal{L}_\text{product}=-\sum_{t=1}^{C}\frac{y_t p_t}{(y_t p_t+{p_t}^2+{y_t}^2)} \enspace,\label{eq:sdmloss}
\end{equation}
\end{small}
where $y_t$ represents the groundtruth SDM and $p_t$ denotes the predicted SDM. The intuition behind taking the product of prediction and groundtruth is that we want to penalize the output SDM for having the wrong sign. In our experiments, we train the SDM prediction model by combining the product loss and $L_1$ loss. Our proposed regression loss based on product is smooth and provides better gradient information when combined with the $L_1$ loss. In Figure~\ref{fig_loss}(b), we compare the combined loss function with the $L_1$ loss. The combined loss focuses more on the values around zero (\emph{i.e.}, boundary represented by SDM) by having larger gradient magnitude. Thus, the combined loss has the potential of improving segmentation result and stabilizing the SDM training.

\begin{table}[b]
\caption{Quantitative comparison of segmentation models on the hippocampus dataset. All models use the same backbone network; SDM denotes training with predicting SDM and our proposed loss; Dice denotes training with predicting segmentation map and the Dice loss. The proposed backbone combining the SDM training and segmentation training achieves best scores in all evaluation metrics.}
\smallskip
\centering
\resizebox{0.99\columnwidth}{!}
{
\smallskip
\begin{tabular}{l|c|c|c|c}
Method & Dice$\uparrow$ & HD(mm)$\downarrow$ & HD95(mm)$\downarrow$ & ASD(mm)$\downarrow$\\
Dice & $0.840\pm0.025$ & $23.568\pm17.811$ & $1.989\pm1.596$ & $0.414\pm0.153$\\
SDM& $0.757\pm0.065$ & $8.076\pm4.650$ & $3.393\pm3.303$ & $0.714\pm0.466$\\
SDM + Dice& $\bm{0.843}\pm0.032$ & $\bm{5.400}\pm1.895$ & $\bm{1.747}\pm1.270$ & $\bm{0.345}\pm0.130$
\end{tabular}
}
\label{table_hippo}
\end{table}

When training SDM and segmentation map jointly, the final loss as shown in Figure~\ref{fig_method} is defined as:
\begin{small}
\begin{equation}
\mathcal{L}=\mathcal{L}_\text{Seg} +  \lambda\mathcal{L}_\text{SDM}=\mathcal{L}_\text{Dice} + \lambda (\mathcal{L}_\text{product} + \mathcal{L}_1)\enspace,\label{eq:finalloss}
\end{equation}
\end{small}
where the final value of $\lambda$ is set to $10$ in all experiments. The value is determined by grid search and experimental results.

\section{Experiments}

\begin{figure}[t]
\centering
\includegraphics[width=0.99\columnwidth]{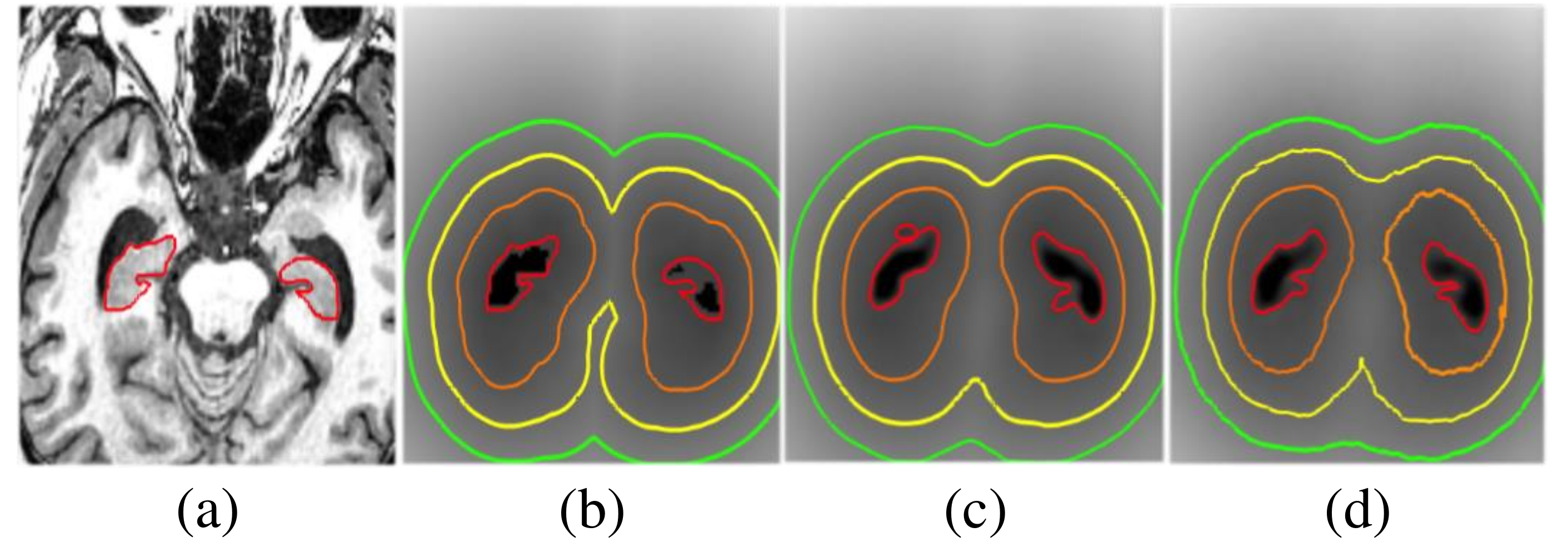} 
\caption{Qualitative SDM comparison on the hippocampus testing set of (a) axial view input image; (b) groundtruth SDM; (c) predicted SDM with SDM training only; (d) predicted SDM from joint training of SDM and segmentation map. From inside to outside, the red, orange, yellow and green contour represents the $0$-, $0.1$-, $0.2$- and $0.3$-distance map, respectively. The grayscale intensity indicates the SDM space. All results are based on the normalized SDM values. 
}
\label{fig_hippo_sdm}
\end{figure}

\begin{figure}[t]
\centering
\includegraphics[width=0.95\columnwidth]{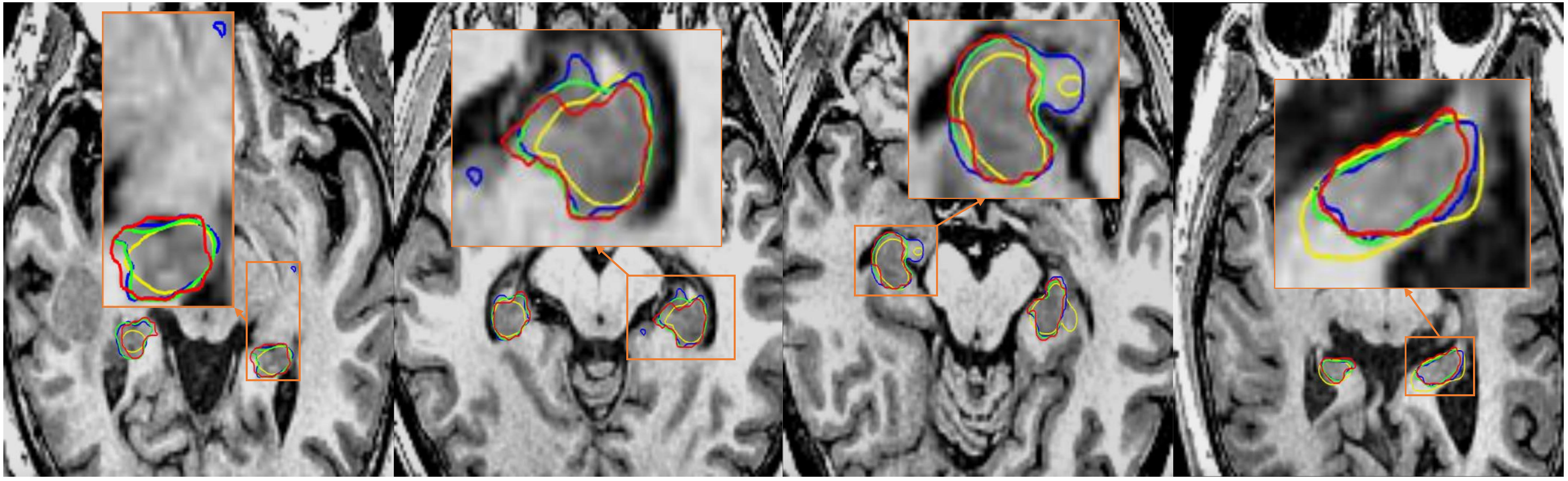} 
\caption{Qualitative segmentation comparison on the hippocampus testing set. The red, blue, yellow and green contours represent the groundtruth annotation, training with segmentation only, training with SDM only, and training with SDM and segmentation jointly. In the first three columns, one can see that the segmentation-only results contain isolated or inaccurate false positive regions. In all columns, the model with joint training achieves best accordance with groundtruth annotation. Zoom in for better view.}
\label{fig_hippo_seg}
\end{figure}

To validate our proposed methods, we conduct comprehensive experiments on a single organ segmentation dataset and a multi-organ segmentation dataset. The single organ dataset is our collected hippocampus segmentation dataset. It contains 72 CT scans from different patients. All scans have unified isotropic spacing of $1$ mm. Groundtruth hippocampus annotations are manually annotated by one experienced doctor in 2D views. See Figure~\ref{fig_hippo_sdm} and Figure~\ref{fig_hippo_seg} for examples of axial view input images and groundtruth annotations. We randomly split the dataset into training set with 60 samples and testing set with 12 samples. All evaluations are performed on the testing set.

We utilize commonly used evaluation metrics for medical image segmentation in all experiments. More specifically, we use Dice coefficient, Hausdorff Distance (HD), 95$\%$ Hausdorff Distance (HD95) and Average Symmetric Surface Distance (ASD) to evaluate different methods. We report the quantitative results of 3 models in Table~\ref{table_hippo}. Using the same backbone 3D segmentation network for all experiments, our proposed jointing training with SDM and segmentation map achieves best performances in all evaluation metrics. Comparing with traditional training with only the segmentation map supervision, our proposed SDM learning method reduces the Hausdorff distance significantly, in both the SDM-only training and the joint training. Since current segmentation algorithms do not capture well the shape information, isolated false positive regions can be generated which results in large HD. Although not affecting the overall result, post-processing such as morphological operations are needed to remove those regions and refine the initial results. On the contrary, SDM training implicitly forces the model to learn the shape information and can greatly reduce such false positives without any post-processing.

We present a group of axial view SDM plots in Figure~\ref{fig_hippo_sdm}. One can observe that the learned SDM with only SDM training obtains smoothest contours, while the joint training of SDM and segmentation map predicts more accurate organ boundary. Overall, they both preserve the shape of hippocampus and align well with the groundtruth SDM. Such results prove that predicting SDM directly from the medical image input is feasible and reliable, where shape information is indeed captured during the learning process. Qualitative comparison of segmentation results is illustrated in Figure~\ref{fig_hippo_seg}. As aforementioned, segmentation results trained with only segmentation output (blue contours) have false positives due to the lack of shape awareness. According to the results shown in Figure~\ref{fig_hippo_seg} and Table~\ref{table_hippo}, segmentation by jointly training with SDM and binary map supervision get best performances. In conclusion, incorporating segmentation with SDM prediction indeed provides meaningful improvements and generates better single organ segmentation results.

\begin{figure*}[t]
\centering
\includegraphics[width=0.95\textwidth]{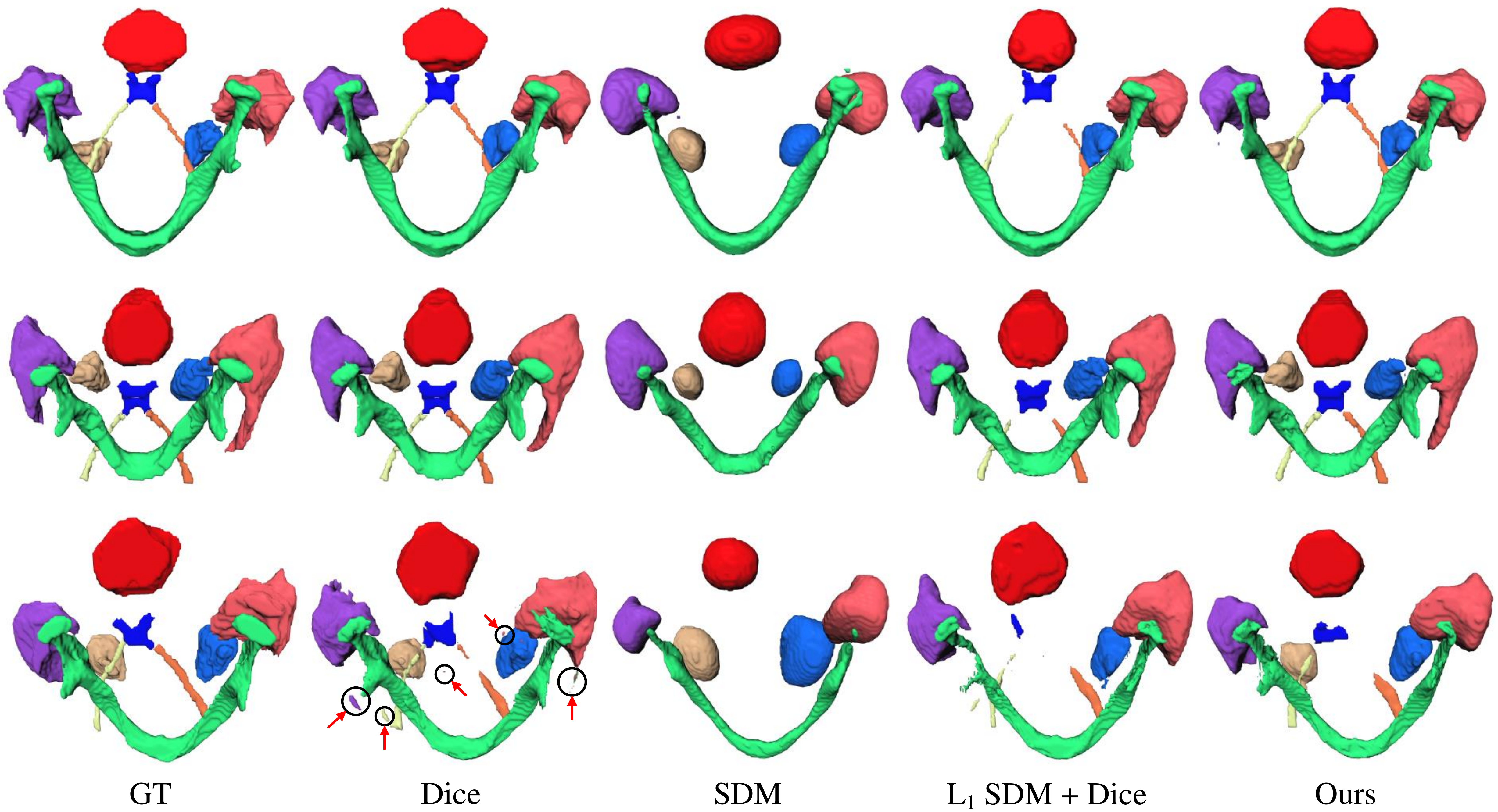} 
\caption{Qualitative results and ablation comparison on the MICCAI 2015 Head and Neck segmentation testing set. Organs are labeled by different colors. First two rows are two samples where both results from Dice only training and the joint training align well with the groundtruth. The third row shows a case where Dice result contains many isolated false positives marked by black circles and red arrows. Our proposed joint training model has clearly smoother and better result in this case.}
\label{fig_pddca_seg}
\end{figure*}

To examine the effectiveness of our proposed method on the more challenging multi-organ segmentation task and compare with current state-of-the-art organ segmentation algorithms, we further conduct experiments on the MICCAI Head and Neck Auto Segmentation Challenge 2015 dataset~\cite{raudaschl2017evaluation}. The MICCAI 2015 dataset is a multi-organ segmentation dataset which contains 38 training CT images and 10 testing images. we crop the head area from the original images since all target organs are inside the head. In Table~\ref{table_pddca_dice} and Table~\ref{table_pddca_hd95}, we compare our proposed methods with other state-of-the-art methods on the same testing set. For both Dice and HD95, our proposed methods improve upon previous state-of-the-arts significantly, especially in small organs such as Chiasm, left and right Optic Nerve. Compared with our backbone network trained with Dice loss only, our joint training model has a slightly lower Dice score, while still outperforms other state-of-the-art methods by a large margin. However, for all other evaluation metrics, the joint training model achieves better scores than training with only segmentation loss.

\begin{table*}[t]
\caption{Dice comparison on the MICCAI 2015 testing set. - indicates the model fails to generate any meaningful segmentation result on that class. With our proposed backbone network, both the segmentation only training and the joint training of SDM and segmentation outperform the current state-of-the-art results by a large margin.}
\smallskip
\centering
\resizebox{0.95\textwidth}{!}
{
\smallskip
\begin{tabular}{l|c|c|c|c|c|c|c}
Organs  & MICCAI2015 & AnatomyNet & FocusNet & Ours & Ours & Ours & Ours \\
~  & \cite{raudaschl2017evaluation} & \cite{zhu2019anatomynet} & \cite{gao2019focusnet} & w/ Dice & w/ SDM & w/ $L_1$+Dice & w/ SDM+Dice \\
\hline
Brain Stem  & $0.880$ & $0.867\pm0.020$ & $0.875\pm0.026$ & $\bm{0.902}\pm0.076$ & $0.765\pm0.072$ & $0.875\pm0.051$ & $0.883\pm0.065$  \\
Chiasm      & $0.557$ & $0.532\pm0.150$ & $0.596\pm0.181$ & $\bm{0.688}\pm0.399$ & - & $0.593\pm0.400$ & $0.658\pm0.364$  \\
Mandible    & $0.930$ & $0.925\pm0.020$ & $0.935\pm0.019$ & $\bm{0.957}\pm0.018$ & $0.812\pm0.045$ & $0.939\pm0.038$ & $0.940\pm0.032$  \\
Opt. Ner. L & $0.644$ & $0.721\pm0.060$ & $0.735\pm0.096$ & $\bm{0.855}\pm0.179$ & - & $0.769\pm0.109$ & $0.841\pm0.147$  \\
Opt. Ner. R & $0.639$ & $0.706\pm0.100$ & $0.744\pm0.072$ & $\bm{0.842}\pm0.192$ & - & $0.763\pm0.126$ & $0.825\pm0.151$  \\
Parotid L   & $0.827$ & $0.881\pm0.020$ & $0.863\pm0.036$ & $\bm{0.893}\pm0.073$ & $0.790\pm0.034$ & $0.875\pm0.059$ & $0.883\pm0.056$  \\
Parotid R   & $0.814$ & $0.874\pm0.040$ & $0.879\pm0.031$ & $\bm{0.888}\pm0.092$ & $0.716\pm0.086$ & $0.867\pm0.087$ & $0.880\pm0.065$  \\
Submand. L  & $0.723$ & $0.852\pm0.040$ & $0.798\pm0.081$ & $\bm{0.857}\pm0.127$ & $0.550\pm0.128$ & $0.826\pm0.110$ & $0.852\pm0.088$  \\
Submand. R  & $0.723$ & $0.813\pm0.040$ & $0.801\pm0.061$ & $\bm{0.847}\pm0.159$ & $0.591\pm0.099$ & - & $0.842\pm0.105$  \\
Average     & $0.723$ & $0.793$ & $0.803$ & $\bm{0.859}$ & - & - & $0.845$ 
\end{tabular}
}
\label{table_pddca_dice}
\end{table*}

\begin{table*}[t]
\caption{$95\%$ Hausdorff (HD95, in mm) comparison on the MICCAI 2015 testing set. Our proposed backbone network with joint training of SDM and segmentation improves the current state-of-the-art by $0.64$ on average.}
\smallskip
\centering
\resizebox{0.95\textwidth}{!}
{
\smallskip
\begin{tabular}{l|c|c|c|c|c|c|c}
Organs  & MICCAI2015 & AnatomyNet & FocusNet & Ours & Ours & Ours & Ours \\
~  & \cite{raudaschl2017evaluation} & \cite{zhu2019anatomynet} & \cite{gao2019focusnet} & w/ Dice & w/ SDM & w/ $L_1$+Dice & w/ SDM+Dice \\
\hline
Brain Stem  & $4.59$ & $6.42\pm2.4$ & $2.14\pm0.6$ & $\bm{1.81}\pm1.10$ & $4.34\pm1.31$ & $2.36\pm0.92$ & $2.29\pm1.48$  \\
Chiasm      & $2.78$ & $5.76\pm2.5$ & $ 3.16\pm1.3$ & $\bm{1.32}\pm1.66$ & - & $2.35\pm3.30$ & $1.37\pm1.55$  \\
Mandible    & $1.97$ & $6.28\pm2.2$ & $1.18\pm0.3$ & $\bm{0.60}\pm0.49$ & $3.14\pm0.86$ & $1.12\pm0.58$ & $1.14\pm0.31$  \\
Opt. Ner. L & $2.76$ & $4.85\pm2.3$ & $ 3.76\pm2.9$ & $11.75\pm31.93$ & - & $6.93\pm2.04$ & $\bm{2.15}\pm3.94$  \\
Opt. Ner. R & $3.15$ & $4.77\pm4.3$ & $2.65\pm1.5$ & $\bm{0.95}\pm1.34$ & - & $5.60\pm2.93$ & $1.32\pm1.99$  \\
Parotid L   & $5.11$ & $9.31\pm3.3$ & $2.52\pm1.0$ & $\bm{1.91}\pm1.16$ & $5.37\pm2.51$ & $2.69\pm1.59$ & $2.32\pm1.32$  \\
Parotid R   & $6.13$ & $10.08\pm5.1$ & $\bm{2.07}\pm0.8$ & $3.27\pm3.66$ & $6.59\pm3.22$ & $3.42\pm3.26$ & $2.46\pm1.56$  \\
Submand. L  & $5.35$ & $ 7.01\pm4.4$ & $2.67\pm1.3$ & $2.54\pm3.03$ & $5.85\pm2.96$ & $\bm{2.27}\pm2.08$ & $2.35\pm2.72$  \\
Submand. R  & $5.42$ & $ 6.02\pm1.8$ & $3.41\pm1.4$ & $5.69\pm11.15$ & $5.08\pm1.68$ & - & $\bm{2.40}\pm1.96$  \\
Average     & $4.14$ & $6.72$ & $2.62$ & $3.32$ & - & - & $\bm{1.98}$ 
\end{tabular}
}
\label{table_pddca_hd95}
\end{table*}

We show qualitative results of ablation study in Figure~\ref{fig_pddca_seg}. Our backbone network trained with only segmentation prediction generally performs well, however, it produces some isolated false positives far away from the actual organ in some cases. The backbone network trained with only SDM prediction has smooth outputs, but does not converge on the small organs including Chiasm, left and right Optical Nerve. The joint training with $L_1$ loss SDM as in~\cite{park2019deepsdf} also fails to get any meaningful segmentation result on right Submandibular. The joint training with our proposed SDM loss converges well on all organs and preserves continuous shape. Except from Dice and HD95 comparison, we also report the HD and ASD comparisons in Table~\ref{table_pddca}. The backbone network trained with our proposed loss achieves best scores in all evaluation metrics. The experimental results prove that the proposed SDM learning algorithm and the new regression loss not only stabilize the training, but also improve the segmentation results.

\begin{table}[t]
\caption{Ablation comparison of our proposed methods on the MICCAI 2015 testing set. The HD (mm) and ASD (mm) are averaged over all organs. Note that the results of SDM only and the joint training with $L_1$ loss is only averaged over organs with meaningful segmentation results.}
\smallskip
\centering
{
\smallskip
\begin{tabular}{l|c|c}
Method & Avg HD$\downarrow$ & Avg ASD$\downarrow$\\
Backbone w/ Dice & $10.95$ & $0.44$ \\
Backbone w/ SDM$^*$& $9.46$ & $1.55$ \\
Backbone w/ $L_1$ + Dice$^*$& $7.47$ & $0.54$ \\
Backbone w/ SDM + Dice& $\bm{5.07}$ & $\bm{0.39}$
\end{tabular}
}
\label{table_pddca}
\end{table}

\subsection{Implementation Details}
In our experiments, we use the same backbone segmentation network for fair comparison. In our proposed deep 3D UNet, the initial number of channels is $24$ and is doubled in each downsampling operation. The maximum number of channels is $384$. All models are trained by Adam optimizer. The initial learning rate is $5e-4$ and decayed by factor of $0.8$ for every $25$ epochs. All models are trained for $200$ epochs for the Hippocampus dataset and $600$ epochs for the MICCAI 2015 dataset. The batchsize is 1 and all experiments are done on a single NVIDIA Tesla P40 GPU with 24G memory. During inference, the segmentation result is obtained by applying the Heaviside function to the predicted SDM.

\begin{figure}[t]
\centering
\includegraphics[width=0.95\columnwidth]{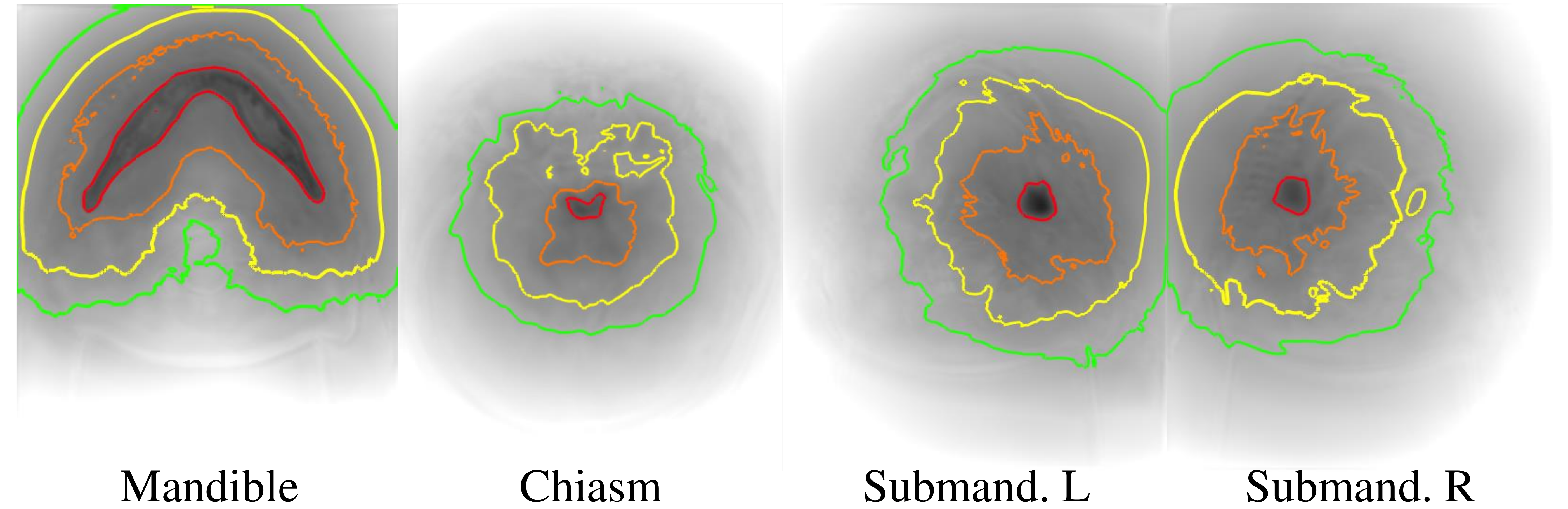} 
\caption{Qualitative SDM prediction results of our proposed joint training of SDM and segmentation map on the MICCAI 2015 test set. The color of contours follows the same rule as in Figure~\ref{fig_hippo_sdm}. We selectively show 4 out of 9 organs as an example. Mandible is with relatively large size; Chiasm is the smallest organ in 9 target organs; Submandibular Left and Right are with medium size.  }
\label{fig_pddca_sdm}
\end{figure}

\section{Discussion}
For the hippocampus segmentation which is a single organ segmentation task, our proposed backbone network with SDM learning achieves superior performances compared to models trained by segmentation loss or SDM loss alone. Moreover, the predicted SDM can produce smooth fixed-distance contours while retaining the shape of hippocampus. For the multi-organ segmentation problem, we show learned SDMs of 4 out of 9 organs in  Figure~\ref{fig_pddca_sdm}. Although still achieving promising segmentation results, the learned SDMs are not ideal SDMs, especially in small organs. Although the overall shape is preserved in Mandible, left and right Submandibular, the shape is not well kept in the Chiasm which is the smallest organ among 9 organs. We assume it indicates that although our method achieves promising segmentation results, the SDMs are not perfectly learned for small organs in the multi-organ segmentation tasks.

Due to the limitation of GPU memory, we use the exact same network architecture for both single organ and multi-organ datasets. In multi-organ segmentation, segmentation maps are predicted in multiple output channels within the same network. However, more information must be propagated through the network in multi-organ segmentation and it naturally requires a network with larger capacity. Currently, one drawback of our multi-SDM model is that, although different organs share the same segmentation network, each organ's SDM is predicted independently in the last layer and there lacks connection between different organs. Since we are not allowed to increase the network capacity accordingly when extending from single organ to multi-organ segmentation, we expect to explore the relationship between the SDMs of different organs to better utilize and share the features learned by convolutional layers in our future work. As we do not set any constrains to input modality, our proposed method can be easily extended to other general semantic or instance segmentation tasks as in~\cite{audebert2019distance}, especially to tasks where object shapes are relatively consistent. Future directions also include better backbone network architecture and training strategy for SDM learning.

\section{Conclusions}

In this work, we shed light on the potential of SDM learning in organ segmentation and explore its advantages over previous segmentation methods. Combining with the traditional segmentation map training, our proposed SDM learning mechanism improves the current state-of-the-art segmentation results by a large margin, both quantitatively and qualitatively. One of the biggest advantages of our SDM learning model is that any existing 3D segmentation network can be easily adapted to incorporate an SDM prediction model with nearly no additional overhead. We believe the SDM learning mechanism has great potentials in various organ segmentation applications including radiotherapy planning and shape analysis, and can be applicable to general segmentation tasks.

\bibliography{arxiv}
\bibliographystyle{aaai}
\end{document}